\documentclass{article}
\usepackage{graphicx} 

\usepackage{times}
\usepackage{soul}
\usepackage{url}
\usepackage[hidelinks]{hyperref}
\usepackage[utf8]{inputenc}
\usepackage[small]{caption}
\usepackage{graphicx}
\usepackage{amsmath}
\usepackage{amssymb}
\usepackage{mathtools}
\usepackage{amsthm}
\usepackage{booktabs}
\usepackage{algorithm}
\usepackage{algorithmic}
\usepackage{multirow}
\usepackage[switch]{lineno}
\usepackage{natbib}
\usepackage{subfigure}
\usepackage{color}
\usepackage{authblk}
\usepackage{xcolor}

\begin{document}

\title{Efficient Hierarchical Implicit Flow Q-learning for Offline Goal-conditioned Reinforcement Learning}

\author[1]{Zhiqiang Dong}
\author[1]{Teng Pang} 
\author[1]{Rongjian Xu}
\author[1 2]{Guoqiang Wu \textsuperscript{*}} 
\affil[*]{{\small Corresponding Author}}
\affil[1]{School of Software, Shandong University, Jinan, Shandong, China \\ zhiqiangdong12@gmail.com, silencept7@gmail.com,
rongjianxu50@gmail.com, guoqiangwu90@gmail.com
}
\affil[2]{Luke EI, Jinan, China
}

\maketitle

\begin{abstract}
   Offline goal-conditioned reinforcement learning (GCRL) is a practical reinforcement learning paradigm that aims to learn goal-conditioned policies from 
reward-free offline data.
Despite recent advances in hierarchical architectures such as HIQL, long-horizon control in offline GCRL remains challenging due to the limited expressiveness of Gaussian policies and the inability of high-level policies to generate effective subgoals.
To address these limitations, we propose the goal-conditioned mean flow policy, which introduces an average velocity field into hierarchical policy modeling for offline GCRL. 
Specifically, the mean flow policy captures complex target distributions for both high-level and low-level policies through a learned average velocity field, enabling efficient action generation via one-step sampling. 
Furthermore, considering the insufficiency of goal representation, we introduce a LeJEPA loss that repels goal representation embeddings during training, thereby encouraging more discriminative representations and improving generalization. 
Experimental results show that our method achieves strong performance across both state-based and pixel-based tasks in the OGBench benchmark.
\end{abstract}

\section{Introduction}
A prevailing trend in modern machine learning is to reduce reliance on domain-specific assumptions while scaling up data.
Offline goal-conditioned reinforcement learning (GCRL) follows this paradigm by training agents to reach arbitrary goal states solely from a fixed, static, and unlabeled (i.e., reward-free) dataset, without interacting with the environment during training~\citep{kaelbling1993learning, levine2020offline, park2024ogbench}.
Despite its promise, offline GCRL remains fundamentally challenging in long-horizon settings.
In particular, accurately estimating goal-conditioned value functions over long horizons is difficult, as naive value learning often leads to high-variance estimates and suboptimal policies~\citep{pateria2021hierarchical, kim2021landmark}.

To mitigate these issues, Hierarchical Implicit Q-Learning (HIQL)~\citep{park2023hiql}, one of the state-of-the-art offline GCRL methods, introduces a hierarchical policy structure in which a high-level policy proposes subgoals and a low-level policy executes primitive actions toward these subgoals.
By decomposing long-horizon decision-making into shorter segments, HIQL provides more stable learning signals for both policy levels, even when the underlying value estimates are noisy.
However, despite its success in certain long-horizon environments, HIQL still exhibits notable limitations.
As offline datasets grow larger and more diverse, their behavior distributions become increasingly complex and multimodal, exposing a key weakness of HIQL: its reliance on unimodal Gaussian policies, which struggle to capture such rich action distributions~\citep{mandlekar2021matters, o2024open}.

Recent advances in generative modeling have inspired new approaches to address policy expressivity in offline RL.
In particular, diffusion models~\citep{ho2020denoising} and flow matching methods~\citep{lipman2024flow} have been applied to offline RL, both for trajectory modeling~\citep{ajay2022conditional, liang2023adaptdiffuser} and policy learning~\citep{wang2022diffusion, chi2025diffusion, park2025flow}.
These generative policies offer significantly greater expressivity than traditional Gaussian or mixture policies.
However, their practical deployment remains challenging: complex multi-step sampling procedures incur substantial computational overhead, and the resulting samples are not directly amenable to standard policy optimization.
To alleviate these limitations, recent work~\citep{kang2023efficient,park2025flow} has explored more efficient training and inference strategies.
Although there has been some progress in these works, they still cannot avoid multi-step sampling.

Motivated by these developments, we propose Hierarchical Implicit Flow Q-Learning (HIFQL), a simple and effective method for offline GCRL.
Inspired by mean flow in generating models~\citep{geng2025mean}, HIFQL extends HIQL by replacing unimodal Gaussian policies with expressive mean flow policies at both the high and low levels.
Specifically, we use mean flow to model the high-level policy that generates subgoals and the low-level policy that produces actions based on these subgoals, enabling expressive yet efficient one-step generation.

Furthermore, in high-dimensional, pixel-based environments, the effectiveness of hierarchical policies critically hinges on learning high-quality goal representations. 
To address this, we incorporate a LeJEPA-based goal representation encoder that learns well-conditioned and semantically meaningful goal embeddings, further improving subgoal prediction and long-horizon performance.

Extensive experiments on the OGBench benchmark demonstrate that HIFQL achieves strong performance across both state-based and pixel-based tasks, highlighting its effectiveness for long-horizon offline goal-conditioned control. Our contributions are summarized as follows:
\begin{itemize}
    \item We propose efficient Hierarchical Implicit Flow Q-Learning (HIFQL), a simple yet effective offline GCRL method that extends hierarchical policy learning by leveraging expressive mean flow models with one-step generation for long-horizon control.
    \item To enhance subgoal prediction in high-dimensional environments, we incorporate a LeJEPA-based goal representation encoder that learns well-conditioned and semantically consistent goal embeddings, improving the robustness of high-level policy learning, especially in pixel-based tasks.
    \item Experiments on the OGBench benchmark demonstrate that HIFQL achieves strong performance on both state- and pixel-based tasks.
\end{itemize}

\section{Preliminaries and Background}
\subsection{Problem setting}
The offline goal-conditioned reinforcement learning (GCRL) problem is defined by a Markov decision process $\mathcal{M}=(\mathcal{S},\mathcal{A},\mu,\mathcal{T}) $, 
where $\mathcal{S}$ denotes the state space, $\mathcal{A}$ denotes the action space, $\mu(\textcolor{gray}{s}) \in \Delta(\mathcal{S})$ denotes the initial state distribution, $\mathcal{T}(\textcolor{gray}{s'\mid s, a}): \mathcal{S} \times \mathcal{A} \to \Delta(\mathcal{S})$ denotes the transition dynamics function. Here $\Delta(\mathcal{X})$ denotes the set of probability distributions defined on a set $\mathcal{X}$ and use \textcolor{gray}{gray} to denote placeholder variables. 
In offline GCRL, the goal is to learn a goal-conditioned policy $\pi(\textcolor{gray}{a} | \textcolor{gray}{s,g}):\mathcal{S} \times \mathcal{S} \to \Delta(\mathcal{A})$ that maximizes the expected return $J(\pi) = \mathbb{E}_{g \sim \mu_{g}(g),\, \tau \sim p^\pi(\tau)} \left[ \sum_{h=0}^{H} \gamma^h r(s_h, g) \right]$,
where $\tau=(s_0, a_0, s_1, \ldots, s_{H})$ denotes the episode, $H \in \mathbb{N}$ denotes the episode horizon, $\gamma \in (0,1)$ denotes the discount factor. Just as in some prior work~\citep{park2023hiql, ghosh2019learning, eysenbach2022contrastive}, we assume that the goal space $\mathcal{G}$ is the same as the state space (i.e., $\mathcal{G}=\mathcal{S})$, $\mu_{g}(\textcolor{gray}{g}) \in \Delta(\mathcal{S})$ denotes a goal distribution, $p^\pi(\tau)$ denotes the trajectory-based distribution given by $p^\pi(\tau) = \mu(s_0) \prod_{h=0}^{H-1} \pi(a_h \mid s_h, g)\, \mathcal{T}(s_{h+1} \mid s_h, a_h)$, and $r(\textcolor{gray}{s,g}):\mathcal{S} \times \mathcal{S} \to \mathbb{R}$ denotes a goal-conditioned reward function\footnote{In this work, we define $r(s,g) = \mathbb{I}(s=g)$, where $\mathbb{I}$ is the indicator function, i.e., if $s=g$, $r(s,g)=1$, else $r(s,g)=0$.}. 
In offline scenarios, we have an unlabeled (i.e., reward-free) dataset $\mathcal{D} = \{\tau^{(n)}\}_{n \in \{1,2,\ldots,N\}}$ consists of $N$ unlabeled trajectories, where each trajectory $\tau^{\{n\}}=(s_0^{(n)}, a_0^{(n)}, s_1^{(n)}, \ldots, s_{H_{n}}^{(n)})$ consists of a sequence of states and actions, and $H_n$ is trajectory-specific horizon.
\begin{figure*}
    \centering
    \includegraphics[width=0.9\linewidth]{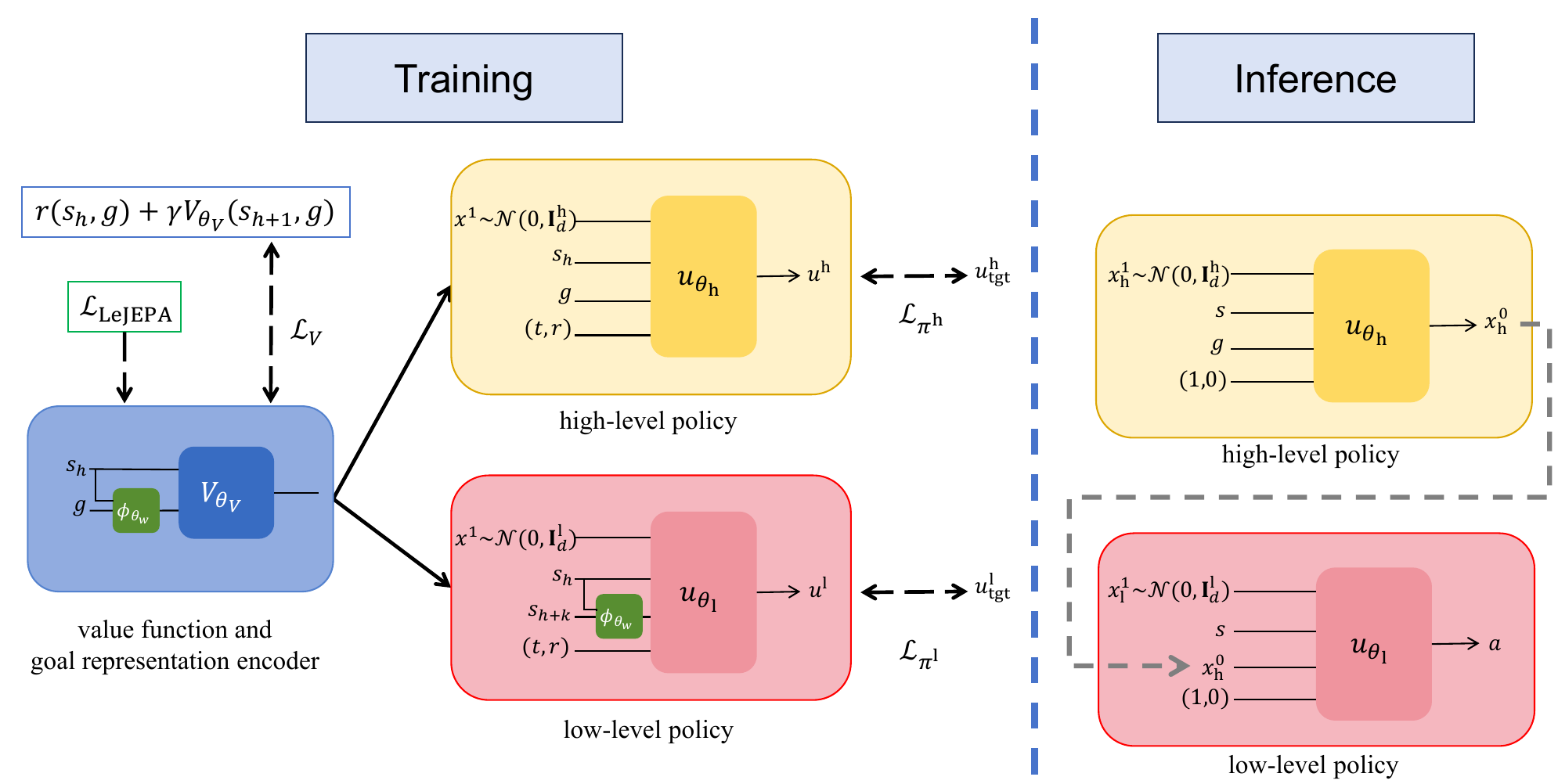}
    \caption{Overview of HIFQL. During training (left), a goal-conditioned value function and a shared goal representation encoder are jointly optimized, while high-level and low-level mean flow policies are trained via advantage-weighted regression to generate subgoals and actions, respectively. During inference (right), both policies perform one-step generation from Gaussian noise, enabling efficient hierarchical decision-making without iterative sampling.}
    \label{fig:overview}
\end{figure*}

\subsection{Hierarchical Implicit Q-Learning(HIQL)}
A key challenge in offline GCRL is long-horizon reasoning, which is the ability to reach a goal state from an initial state via a long sequence of step~\citep{huang2019mapping,kim2021landmark,park2024ogbench}.
To tackle this issue, hierarchical implicit Q-learning~(HIQL)~\citep{park2023hiql} adopts a hierarchical modeling approach for training and execution. It has three components: a high-level policy generating subgoals, a low-level policy producing actions, and a goal-conditioned value function trained with implicit Q-learning~\citep{kostrikov2021offline}.
Specifically, HIQL trains a goal-conditioned value function $V_{\theta_V}(s,g)$ with the following loss:

\begin{align}
    \label{value}
    \mathcal{L}_V(\theta_V) =
    \mathbb{E}_{\substack{(s_h,s_{h+1}) \sim \mathcal{D},\\ g \sim \mu_g(g)}}  \big[ L_2^\kappa  \big(r(s_h,g) + \gamma V_{\bar{\theta}_V }(s_{h+1},g) 
     - V_{\theta_V}(s_h,g)\big)\big] ,
\end{align}
where $\bar{\theta}_{V}$ denotes the parameters of the target goal-conditioned value function network, and $L_2^\kappa$ is the expectile loss with a parameter $\kappa  \in [0.5,1): L_2^\kappa (x)=|\kappa  - \mathbb{I}(x<0)|x^2$. For policy learning, both high-level and low-level policies are extracted using advantage-weighted regression (AWR)~\citep{peng2019advantage},
optimized according to the following objective:
\begin{equation}\label{eq:HIQL}
\begin{aligned}
    \mathcal{L}_{\mathrm{\pi^h}}(\theta_\mathrm{h}) &= 
    \mathbb{E}_{\substack{(s_h,s_{h+k}) \sim \mathcal{D},\\ g \sim \mu_g(g)}}
    \big [ \mathrm{exp}\big(\beta A^\mathrm{h}(s_h,s_{h+k},g) \big)  
    \times \log\pi_{\theta_\mathrm{h}}^\mathrm{h}(s_{h+k} \mid s_h,g) \big], \\
    \mathcal{L}_{\pi^\mathrm{l}}(\theta_\mathrm{l}) &= 
    \mathbb{E}_{(s_h,a_h,s_{h+1}, s_{h+k})\sim \mathcal{D}}
    \big [\mathrm{exp}\big(\beta A^\mathrm{l}(s_h,s_{h+1},s_{h+k})\big)   
    \times \log\pi_{\theta_\mathrm{l}}^\mathrm{l}(a_h \mid s_h,s_{h+k}) \big ] , 
\end{aligned}
\end{equation}

where $\beta$ denotes the inverse temperature, $s_{h+k}$ represents the optimal $k$-step subgoal with $k \in \mathbb{N}$,
$A^\mathrm{h}(s_{h},s_{h+k},g)=V_{\theta_\mathrm{V}}(s_{h+k},g)-V_{\theta_\mathrm{V}}(s_h,g)$ and $A^\mathrm{l}(s_{h},s_{h+1},s_{h+k})=V_{\theta_\mathrm{V}}(s_{h+1}, s_{h+k})-V_{\theta_\mathrm{V}}(s_h,s_{h+k})$ denote the advantage functions of the high-level policy $\pi^\mathrm{h}$ and low-level policy $\pi^\mathrm{l}$, respectively.
To address the infeasibility of subgoals prediction in high-dimensional domains, HIQL incorporates representation learning and presents a goal representation function $\phi_{\theta_\mathrm{w}}(s,g)$. 
This hierarchical structure mitigates the signal-to-noise ratio challenge, allowing the agent to produce reliable actions, particularly for long-horizon tasks where the goal is far away.
Nevertheless, HIQL faces two unresolved challenges: 1) The Gaussian policy is inherently incapable of effectively modeling complex, multi-modal target distributions, which demands a more expressive policy; 2) The goal representation function lacks the capacity to extract adequate and useful representations from high-dimensional domains.

\subsection{Flow Matching}
\label{author info}

As a simpler generative paradigm for continuous-time normalization flows~(CNF), flow matching~\citep{lipman2022flow,liu2022flow, albergo2022building} enables efficient sampling via deterministic ordinary differential equations (ODEs). 
Unlike diffusion-based approaches that rely on stochastic dynamics, flow matching formulates generation as a deterministic transport process, which substantially simplifies both training and sampling.
Formally, given a data distribution $p_{\mathrm{data}}(\textcolor{gray}{x}) \in \Delta(\mathbb{R}^d)$ over a $d$-dimensional Euclidean space, flow matching aims to train a neural network $v_\theta(\textcolor{gray}{t,x}):[0,1] \times \mathbb{R}^d \rightarrow \mathbb{R}^d$ to parameterize a time-dependent velocity field $v(t,x)$.
The corresponding flow $\psi(\textcolor{gray}{t,x}):[0,1] \times \mathbb{R}^d \rightarrow \mathbb{R}^d$ of this velocity field is the unique solution to the ODE:
\begin{align*}
    \frac{d}{dt}\psi(t,x) = v_\theta(\psi(t,x)), \notag
\end{align*}
This flow deterministically transforms samples from a simple base distribution
$q(\textcolor{gray}{x}) \in \Delta(\mathbb{R}^d)$ (e.g., a standard Gaussian $\mathcal{N}(0, \mathbf{I}_d)$) at $t=1$
into samples from the target data distribution $p_{\mathrm{data}}(\textcolor{gray}{x})$ at $t=0$.

In this work, we focus on the simplest formulation of flow matching, which employs linear interpolation paths and uniform sampling over time~\citep{lipman2024flow}. The flow matching loss is as follows:
\begin{align*}
    \mathcal{L}_{\mathrm{FM}}(\theta) = \mathbb{E}_{\substack{
    x^0 \sim p_{\mathrm{data}}(\textcolor{gray}{x}),\\
    x^1\sim q(\textcolor{gray}{x}),\\
    t \sim \mathrm{Unif}([0,1])
    }}\left\|v_\theta(t,x^t)-(x^1-x^0)\right\|_2^2, \notag
\end{align*}
where $\mathrm{Unif}([0,1])$ denotes a uniform probability distribution supported on the interval $[0,1]$, and $x^t=tx^1 + (1-t)x^0$ is the linear interpolation between $x^0$ and $x^1$. For the training phase, the objective learns a vector field whose induced dynamics recover the data distribution $p_{\mathrm{data}}$. At inference time, samples are obtained by numerically solving the ODE specified by 
$v_\theta$.

\section{Method}

In this section, we present \textbf{Hierarchical Implicit Flow Q-Learning (HIFQL)}~(in Figure \ref{fig:overview}), a unified method for offline goal-conditioned reinforcement learning (GCRL).
Our method extends HIQL by replacing unimodal Gaussian policies with expressive mean flow policies at both the high level and low level, enabling efficient one-step generation.
We further introduce a LeJEPA-based goal representation encoder to improve subgoal prediction, particularly in high-dimensional, pixel-based environments.
Together, these components yield an effective and scalable solution for long-horizon offline goal-conditioned control.
\subsection{Hierarchical Goal-conditioned Mean Flow Policy}

HIQL~\citep{park2023hiql} provides reliable actions for long-horizon goals, but its Gaussian-based modeling approach constrains the expressive capacity of the two-level architecture.
Based on this, we extend the modeling approach of HIQL to flow-based generative models.
Following HIQL, our approach adopts a two-level architecture overall: a high-level policy that proposes subgoals (i.e., $s_{h+k}$) and a low-level policy that executes primitive actions $a$ toward generated subgoals $s_{h+k}$.
For simplicity, we use $\pi^{g}(x^0|s_h,y)$ to denote both policies simultaneously: when the goal $y=g$, $\pi^{g}$ represents the high-level policy with $x^0=s_{h+k}$, otherwise $\pi^{g}$ denotes the low-level policy with $x^0=a_h$.
Similar to previous work~\citep{park2025flow}, we define the flow-based goal-conditioned policies with the instantaneous velocity field $v^g(x^t, t, s_h, y)$, where $x^t$ can be obtained by a linear interpolation between the noise $x^1 \sim \mathcal{N}(0, \mathbf{I}_d^h)$ and the target $x^0$. 
This velocity field transforms noise into the target action $a_h$ or subgoal $s_{h+k}$ via numerical integration, conditioned on the state $s_h$ and goal $y$.
Despite using the instantaneous velocity can achieve high expressiveness, this approach suffer from significant inference latency due to multi-step sampling.


To overcome this limitation, following mean flow~\citep{geng2025mean}, we adopt mean flow policies, which learn the average velocity field over a time interval $[r,t]$, enabling one-step generation at inference.
Formally, the average velocity field is defined as
\begin{align*}
    u^g(x^t, r, t, s_{h}, y) \triangleq
    \frac{1}{t-r}
    \int_r^t v^g(x^\tau, \tau, s_h, y)\,\mathrm{d}\tau.
\end{align*}

It is intractable to obtain the average velocity field $u^g$ directly from above definition due to the integral term. 
Following mean flow, we decompose the above definition.
Firstly, we multiply both sides by $(t-r)$, and then differentiate both sides with respect to $t$, treating $r$ as independent of $t$. This leads to:

\begin{align}\label{meanflow_expansion1}
    u^g(x^t,r,t,s_h,y) = v^g(x^t,t,s_h,y)  
    - (t-r)\frac{d}{dt}u^g(x^t,r,t,s_h,y).
\end{align}

For our hierarchical goal-conditioned mean flow policy, we use a shared goal encoder
$\phi_{\theta_\omega}$ to obtain goal representations, i.e., $x^0=\phi_{\theta_{\omega}}(s_h,s_{h+k})$ when $y=g$.
Mean flow is then employed to model both the high-level policy, which generates subgoal representations, and the low-level policy, which produces primitive actions conditioned on these subgoals.

\paragraph{High-level mean flow policy.}
We set $y=g$ and train the parameterized high-level policy $u_{\theta_\mathrm{h}}$ to generate subgoal representations $\phi_{\theta_{\omega}}(s_h,s_{h+k})$ via mean flow regression:
\begin{align}
\label{meanflowloss}
\mathcal{L}_{\mathrm{HMF}}(\theta_\mathrm{h})=
\mathbb{E}\Big[\big\|u_{\theta_{\mathrm{h}}}(x^t, r, t, s_{h}, y)-\mathrm{sg}(u_{\mathrm{tgt}}^{\mathrm{h}})\big\|_2^2\Big],
\end{align}
where $(s_{h},s_{h+k}) \sim \mathcal{D}$, $g \sim \mu_{g}(g)$, $x^0=\phi_{\theta_{\omega}}(s_h,s_{h+k})$, $x^1 \sim \mathcal{N}(0, \mathbf{I}_d^\mathrm{h})$, $(r,t) \sim \mathrm{Unif}([0,1])$ with $r<t$ and $\mathrm{sg}(\cdot)$ denotes the stop-gradient operator for training stability.
The target average velocity is defined using Eq.~\eqref{meanflow_expansion1}:
\begin{align} \label{u_tgt_h}
u_{\mathrm{tgt}}^{\mathrm{h}} =v^{\mathrm{h}}-(t-r)\frac{d}{dt}u_{\theta_{\mathrm{h}}}(x^t, r, t, s_{h}, y),
\end{align}
where $v^{\mathrm{h}}=x^1-x^0=x^1-\phi_{\theta_\omega}(s_h,s_{h+k})$.

\paragraph{Low-level mean flow policy.}
Similarly, we set $y=\phi_{\theta_{\omega}}(s_h,s_{h+k})$ and train the parameterized low-level policy $u_{\theta_\mathrm{l}}$ to generate action $a_{h}$:
\begin{align}
\label{meanflowlow}
\mathcal{L}_{\mathrm{LMF}}(\theta_\mathrm{l})=\mathbb{E}\Big[ \big\| u_{\theta_{\mathrm{l}}}(x^t, r, t, s_h, y) - \mathrm{sg}(u_{\mathrm{tgt}}^{\mathrm{l}}) \big\|_2^2 \Big],
\end{align}
where $(s_{h},a_h,s_{h+k})\sim\mathcal{D}$, $x^0=a_h$ and $x^1 \sim \mathcal{N}(0, \mathbf{I}_d^\mathrm{l})$.
The target average velocity is
\begin{align}\label{u_tgt_l}
u_{\mathrm{tgt}}^{\mathrm{l}} &= v^{\mathrm{l}} - (t-r)\frac{d}{dt}u_{\theta_\mathrm{l}}(x^t, r, t, s_h, y), 
\end{align}
where $v^{\mathrm{l}}=x^1-x^0=x^1-a_h$.

In Eq.~\eqref{meanflow_expansion1}, the total derivative $\frac{d}{dt}u^g(x^t,r,t,s_h,y)$ is computed efficiently using Jacobian-vector product (JVP). 
Therefore, we can calculate $u^{\mathrm{h}}_{\mathrm{tgt}}$ and $u^{\mathrm{l}}_{\mathrm{tgt}}$ through JVP and then train the two-level mean flow policies by minimizing Eq.~\eqref{meanflowloss} and Eq.~\eqref{meanflowlow}.

\subsection{LeJEPA Goal Representation Encoder}

Recent work~\citep{ahn2025option} shows that while HIQL can accurately execute low-level policies given short-horizon subgoals, its main challenge lies in handling the inaccuracies of the high-level policy in long-horizon scenarios.
The effectiveness of the high-level policy critically depends on goal representation, particularly in pixel-based environments, where inadequate or poorly structured representations severely degrade subgoal prediction.

To directly address this issue and effectively extract the necessary information from the original states and goals, we introduce a goal representation encoder based on LeJEPA~\citep{balestriero2025lejepa}. 
LeJEPA is a specialized Joint-Embedding Predictive Architectures (JEPA)~\citep{bromley1993signature,lecun2022path,monemi2025tutorial} that captures semantic relationships and predicts abstractions through latent embeddings
Unlike conventional JEPA approaches that require manually designed architectural constraints~\citep{chen2020simple, chen2021empirical}, LeJEPA verifies that the isotropic Gaussian distribution is the optimal target distribution for JEPA embeddings and introduces an objective function, Sketch Isotropic Gaussian Regularization (SIGReg), which constrains the learned embeddings toward the desired ideal distribution.

Following LeJEPA, we first need to build a usable latent embedding that captures semantic relationships.
Let $z_{n}^\nu$ denote the latent goal embedding corresponding to the $\nu$-th view of the $n$-th sample, obtained by applying the shared goal encoder $\phi_{\theta_{\omega}}(\cdot,\cdot)$ to different transformations of the state-goal pair $(s,g)$. 
Considering that different views should capture complementary semantic information, including temporal consistency and goal invariance, which are crucial for stable high-level policy learning in offline GCRL, we employ four distinct views, defined as follows:
\begin{align}
    z_{n}^1=\phi([s_h,g]), \quad
    &z_{n}^2=\phi([s_{h+k},g]),  \notag \\
    z_{n}^3=\phi([s_{h}^{\mathrm{aug}},g]), \quad
    &z_{n}^4=\phi([s_h,g^{\mathrm{aug}}]).   \notag
\end{align}
Here, $z_{n}^1$ corresponds to the current state-goal pair, $z_{n}^2$ uses a future state $s_{h+k}$ sampled from the same trajectory to encode temporal consistency, 
$z_{n}^3$ and $z_{n}^4$ are visual-augmented views constructed based on the state $s_{h}$ and the goal $g$.
To encourage the above views to capture consistent semantic information in the latent space, the following prediction loss is introduced to enforce consistency among their latent representations explicitly:
\begin{align}
    \mathcal{L}_{\mathrm{pred}} (\{{z_{n}^\nu}\}_{\nu=1}^V)&= \frac{1}{V} \sum_{\nu=1}^{V} 
    \left\| 
    z_{n}^\nu- \bar{z}_{n}
    \right\| ^2_2 ,  \notag 
\end{align}
where $\bar{z}_{n} = \frac{1}{V} \sum_{\nu=1}^{V}z_{n}^\nu$.

While the prediction loss enforces semantic consistency across different views of the same state–goal pair, it does not impose any constraint on the global distribution or geometry of the embedding space. 
As a result, the learned representations may still suffer from poor conditioning or degenerate distributions, which can harm generalization and downstream policy learning. 
To address this limitation, we further regularize the embedding distribution using SIGReg:
\begin{align}
    \mathrm{SIGReg}\!\left(\{z_{n}^\nu \}_{n=1}^{N}\right)
    =\frac{1}{M}\sum_{m=1}^{M}N \notag 
    \int\left|\hat{\varphi}_{\nu,m}(\omega)-\varphi_{\mathcal{N}}(\omega)\right|^{2}\,\varphi_{\mathcal{N}}(\omega)\,\mathrm{d}\omega ,\notag
\end{align}
where $M$ denotes the number of random projections (sketches) used in SIGReg,
and $\omega \in \mathbb{R}$ is the Fourier-domain variable.
For the $m$-th projection, $\hat{\varphi}_{\nu,m}(\omega)$ denotes the empirical
characteristic function of the projected embeddings: 
\begin{align}
    \hat{\varphi}_{\nu,m}(\omega)
    = \frac{1}{N}\sum_{n=1}^{N}
    \exp\!\left(i\,\omega\, a_m^\top z_{n}^\nu \right), \notag
\end{align}
where $a_m \in \mathbb{R}^d$ is a unit-norm random projection vector.
$\varphi_{\mathcal{N}}(\omega)=\exp(-\omega^2 / 2\sigma^2)$ denotes the
characteristic function of an isotropic Gaussian distribution
$\mathcal{N}(0, \mathbf{I}_d^\mathrm{h})$.

The overall LeJEPA objective is defined as a weighted combination of the
prediction loss and the SIGReg: 
\begin{equation}
    \label{LE}
\begin{aligned}
     \mathcal{L}_\mathrm{LeJEPA}(\{z_{n}^\nu\}_{n,\nu=1}^{N,V}) = \frac{\alpha}{V}\sum_{\nu=1}^V \mathrm{SIGReg}(\{\{z_{n}^\nu\}_{n=1}^N\})
    + \frac{1-\alpha}{N}\sum_{n=1}^N \mathcal{L}_{\mathrm{pred}}(\{ z_{n}^\nu \}_{\nu=1}^V). 
\end{aligned}
\end{equation}

where $\alpha$ denotes the weighting coefficient.

\subsection{Training Objective and Inference}

In offline GCRL, relying solely on behavior cloning is insufficient for effective exploration and utilization of existing behaviors. Therefore, similar to HIQL, we incorporate the value function as an evaluation signal for the two-level architecture and optimize this hierarchical policy architecture using weighted behavior cloning~\citep{peng2019advantage}.
Specifically, we jointly train the value function $V_{\theta_V}$ and the goal representation encoder $\phi_{\theta_\omega}$ end-to-end, and optimize the hierarchical policy with a weighted mean flow objective.


\paragraph{Critic and goal representation learning.}
The goal-conditioned value function $V_{\theta_V}(s_h,\phi_{\theta_\omega}(s_h,g))$ is trained using expectile regression with Eq.\eqref{value}.
Concurrently, the goal representation encoder $\phi_{\theta_\omega}(s_h,g)$ is regularized using the LeJEPA loss to enforce both semantic consistency across views and well-conditioned latent geometry.

The overall critic objective is given by
\begin{align}
    \label{critic}
    \mathcal{L}_{\mathrm{critic}}(\theta_V,\theta_w) = \mathcal{L}_{V}(\theta_V,\theta_w) + \lambda\mathcal{L}_{\mathrm{LeJEPA}}(\theta_w),
\end{align}
where $\lambda$ is the representation regularization coefficient.
\paragraph{Policy optimization.}
We train both the high-level policy $u_{\theta_{\mathrm{h}}}$ and low-level policy $u_{\theta_{\mathrm{l}}}$ using the weighted mean flow objective based on goal-conditioned value function $V_{\theta_V}$.

Specifically, the high-level policy is trained to generate subgoals $\phi_{\theta_{\omega}}(s_h,s_{h+k})$ by minimizing:
\begin{align}
    \label{highp}
    \mathcal{L}_{\pi^{\mathrm{h}}}(\theta_\mathrm{h})&=
    \mathbb{E}_{\substack{x^1 \sim \mathcal{N}(0,\mathbf{I}_d^\mathrm{h}) ,\\ (t,r) \sim \mathrm{Unif}([0,1]), r<t,  \\(s_h,s_{h+k}) \sim \mathcal{D},\\g \sim \mu_{g}(g)}}
    \Big[ \exp\Big(\beta A^{\mathrm{h}}(s_h,s_{h+k},g)\Big)  \notag \\ 
    & \quad \times \left\| u_{\theta_{\mathrm{h}}}(x^t, r, t, s_h, g) - \mathrm{sg}(u_{\mathrm{tgt}}^\mathrm{h}) \right\|^2_2 \Big],
\end{align}
where $x^0=\phi_{\theta_{\omega}}(s_h,s_{h+k})$, the target mean velocity $u_{\mathrm{tgt}}^\mathrm{h}$ can be obtained with Eq.\eqref{u_tgt_h}.

Similarly, the low-level policy is trained to generate primitive actions $a_h$ with the current state $s_h$ and predicted subgoal representation $\phi_{\theta_{\omega}}(s_h,s_{h+k})$:
\begin{align}
\label{lowp}
    \mathcal{L}&_{\pi^{\mathrm{l}}}(\theta_\mathrm{l})=
    \mathbb{E}_{\substack{x^1 \sim \mathcal{N}(0,\mathbf{I}_d^\mathrm{l}), \\(t,r) \sim \mathrm{Unif}([0,1]), r<t ,\\ (s_h,a_h,s_{h+1}s_{h+k}) \sim \mathcal{D},\\g \sim \mu_{g}(g)}} 
    \Big[ \exp \Big(\beta A^{\mathrm{l}}(s_h, s_{h+1}, s_{h+k})\Big)\notag\\ & \quad \times \left\| u_{\theta_{\mathrm{l}}}(x^t, r, t, s_h, \phi_\omega([s_h,s_{h+k}])) - \mathrm{sg}(u_{\mathrm{tgt}}^\mathrm{l}) \right\|^2_2 \Big],
\end{align}
where $x^0=a_h$, the target mean velocity $u_{\mathrm{tgt}}^\mathrm{l}$ can be obtained with Eq.\eqref{u_tgt_l}.

\begin{algorithm}
\caption{HIFQL: Training}
\label{ALGO}
\begin{algorithmic}[1] 
    \renewcommand{\algorithmicrequire}{\textbf{Input:}}
    \renewcommand{\algorithmicensure}{\textbf{Output:}}

    \STATE \textbf{Input}: Offline goal-conditioned dataset $\mathcal{D}$; value function $V_{\theta_V}$, the goal representation encoder $\phi_{\theta_w}$, high-level average velocity model~ $u_{\theta_\mathrm{h}}$ and low-level average velocity model~$u_{\theta_\mathrm{l}}$; 
    learning rates $\lambda_V, \lambda_w, \lambda_\mathrm{h}, \lambda_\mathrm{l}$, 
    LeJEPA weight coefficient $\alpha$ and representation regularization coefficient $\lambda$
    \STATE Initialize $V_{\theta_V}$, $\phi_{\theta_w}$, $u_{\theta_\mathrm{h}}$ and $u_{\theta_\mathrm{h}}$
    \WHILE{not converged}
    \STATE  Sample batch $\{{(s_h,a_h,s_{h+k})^{(n)}\}_{n=1}^N}\sim \mathcal{D}, \{g^{(n)}\}_{n=1}^N\sim \mu_g(g)$
    \STATE  $\theta_V\leftarrow\theta_V - \lambda_V \nabla_{\theta_V} \mathcal{L}_{\mathrm{critic}}(\theta_V)$, 
    $\theta_w\leftarrow\theta_w - \lambda_w \nabla_{\theta_w} \mathcal{L}_{\mathrm{critic}}(\theta_w)$ with Eq.~\eqref{critic}
    \STATE  $\theta_\mathrm{h}\leftarrow\theta_\mathrm{h} - \lambda_\mathrm{h} \nabla_{\theta_\mathrm{h}} \mathcal{L}_{\pi^\mathrm{h}}(\theta_\mathrm{h})$ with Eq.~\eqref{highp}
    \STATE  $\theta_\mathrm{l}\leftarrow\theta_\mathrm{l} - \lambda_\mathrm{l} \nabla_{\theta_\mathrm{l}} \mathcal{L}_{\pi^\mathrm{l}}(\theta_\mathrm{l})$ with Eq.~\eqref{lowp}
    \ENDWHILE

\end{algorithmic}
\end{algorithm}

\begin{algorithm}
\caption{HIFQL: Inference}
\label{ALGO:Sampling}
\begin{algorithmic}[1] 
    \renewcommand{\algorithmicrequire}{\textbf{Input:}}
    \renewcommand{\algorithmicensure}{\textbf{Output:}}

    \STATE \textbf{Input}: state $s$, goal $g$

    \STATE Sample $x^{1}_\mathrm{h} \sim \mathcal{N}(0,\mathbf{I}_d^\mathrm{h})$
    \STATE Compute $x^0_\mathrm{h} = x_\mathrm{h}^1 - u_{\theta_\mathrm{h}}(x_\mathrm{h}^1,0,1,s,g)$
    \STATE Sample $x^1_\mathrm{l} \sim \mathcal{N}(0,\mathbf{I}_d^\mathrm{l})$
    \STATE Compute $a = x_\mathrm{l}^1 - u_{\theta_\mathrm{l}}(x_\mathrm{l}^1,0,1,s, x^0_\mathrm{h})$

\end{algorithmic}
\end{algorithm}

\paragraph{Inference.}
For the inference phase, HIFQL performs hierarchical decision-making in a fully one-step manner.
Given a state-goal pair $(s,g)$, the high-level mean flow policy generates a subgoal representation $x^0_{\mathrm{h}}$ in one step from Gaussian noise $x^1_{\mathrm{h}} \sim \mathcal{N}(0, \mathbf{I}_d^\mathrm{h})$:
\begin{align}\label{sample:high}
   x^0_{\mathrm{h}} = x_\mathrm{h}^1 - u_{\theta_\mathrm{h}}(x_\mathrm{h}^1,0,1,s,g).
\end{align}
Conditioned on this subgoal, the low-level mean flow policy produces the corresponding action $a$ with Gaussian noise $x^1_{\mathrm{l}} \sim \mathcal{N}(0, \mathbf{I}_d^\mathrm{l})$ in the same way:
\begin{align}\label{sample:low}
    a = x_\mathrm{l}^1 - u_{\theta_\mathrm{l}}(x_\mathrm{l}^1,0,1,s, x^0_{\mathrm{h}}).
\end{align}
This design eliminates iterative sampling at inference and enables efficient deployment in long-horizon settings.

We present the complete training and inference processes of the HIFQL in Algorithms~\ref{ALGO} and Algorithm~\ref{ALGO:Sampling}.

\section{Related Work}
\subsection{Offline Goal-conditioned Reinforcement Learning}
Offline goal-conditioned reinforcement learning~(GCRL) is a natural analogy to data-driven self-supervised learning within the field of reinforcement learning, which aims to learn a universal goal-conditioned policy capable of navigating from any arbitrary initial state to any target state within the unlabeled (i.e., reward-free) dataset via the fewest number of steps~\citep{kaelbling1993learning, liu2022goal, park2024ogbench}. Existing methods in offline GCRL typically build on metric learning~\citep{wang2023optimal,myers2024learning}, which learns distance functions to measure state or goal similarity, contrastive learning~\citep{ma2022vip, eysenbach2022contrastive, liu2024single}, which encourages representations that distinguish positive state-goals pairs from negative ones, and hierarchical policy learning~\citep{park2023hiql}, which decomposes decision-making into multiple levels of abstraction, enabling high-level policies to guide ow-level action execution. Despite their advancements, these methods often rely on unimodal Gaussian policies, which fail to capture the complex, multi-modal action distributions inherent in offline goal-conditioned datasets. To address this limitation, we introduce HIFQL, a method specifically designed to model these complex distributions using a highly expressive mean flow policy while also incorporating an improved goal representation to extract richer information from the original data.
\subsection{Diffusion-based and Flow-based Reinforcement Learning}
Inspired by the recent successes of generative models such as denoising diffusion models~\citep{sohl2015deep, ho2020denoising, dhariwal2021diffusion}, and flow matching~\citep{lipman2022flow, esser2024scaling}, a variety of approaches have been developed to integrate these techniques into reinforcement learning effectively. Diffuser~\citep{janner2022planning}, Decision Diffuser~\citep{ajay2022conditional}, and AdaptDiffuser~\citep{liang2023adaptdiffuser} applied generative models to model trajectories. DiffusionQL~\citep{wang2022diffusion} uses generative models to model policy. While these methods demonstrate superior expressivity, their reliance on multi-step sampling incurs a prohibitive computational overhead. This latency bottleneck is particularly acute in policy learning, where the algorithm demands rapid and repeated trajectory rollouts. To improve the sample efficiency of policy learning in diffusion and flow models, EDP~\citep{kang2023efficient} leverages a value-weighted diffusion training paradigm, while FQL~\citep{park2025flow} distills the policy into an efficient one-step policy. In the field of image generation, recent research has started to explore methods for training one-step generative flow models~\citep{song2023consistency,lu2024simplifying,geng2024consistency}. Mean flow, as a representative approach, achieves the strongest one-step generation on ImageNet~\citep{geng2025mean}. We propose a goal-conditioned policy with high expressive capacity that incorporates mean flow to model the policy and enables one-step target generation.

\section{Experiments}
\subsection{Experiment setup}
\textbf{Benchmark.}
\quad
We evaluate our method on OGBench~\citep{park2024ogbench}, a recently introduced offline goal-conditioned reinforcement learning~(GCRL) benchmark designed to assess long-horizon decision-making with specified goals.

OGBench comprises a suite of Maze-based control tasks that pose challenging long-horizon planning problems under sparse reward supervision. In these tasks, an agent must navigate from an initial state to a specified goal location, potentially requiring non-trivial exploration, temporal abstraction, and trajectory composition. The Maze environments are categorized by agent type, including PointMaze (2-DoF), AntMaze (8-DoF), and HumanoidMaze (21-DoF), with increasing state and action dimensionality.

Beyond standard navigation tasks, OGBench also includes stitching tasks, which require the agent to compose behaviors across disconnected trajectory segments, further amplifying the difficulty of long-horizon reasoning in offline settings. Following the OGBench evaluation protocol, we consider a total of 13 state-based sparse-reward tasks spanning both navigation and stitching scenarios, as well as 4 pixel-based sparse-reward navigation tasks.

\begin{table*}[t]
\centering
\setlength{\tabcolsep}{6pt}
\renewcommand{\arraystretch}{1.15}
\caption{Results on state-based tasks. We evaluate all methods on three Maze environment families—PointMaze, AntMaze, and HumanoidMaze—covering navigation, stitching, and teleport variants with increasing state dimensionality and horizon length. A total of 13 sparse-reward tasks are included. We report the average success rate and standard deviation over 5 random seeds for each algorithm. The best result for each task is shown in \textbf{bold}.}
\label{tab:maze_results}
\resizebox{\textwidth}{!}{
\begin{tabular}{lllcccccc}
\toprule
Environment & Dataset & GCBC & GCIVL & GCIQL & QRL & CRL & HIQL & HIFQL (ours)\\
\midrule
\multirow{7}{*}{\textbf{pointmaze}} 
& pointmaze-medium-navigate-v0  & $9.8 \pm 6.7$  & $73.8 \pm 5.6$ & $53.5 \pm 3.9$ & $ 86.2\pm 4.2$ & $34.7 \pm 5.6$ & $74.6 \pm 3.3$ & \textbf{99.5} $\pm$ \textbf{0.9} \\
& pointmaze-large-navigate-v0   & $27.3 \pm 7.0$ & $49.8 \pm 8.0$ & $31.6 \pm 3.2$ & $85.8 \pm 10.1$ & $35.5 \pm 13.3$ & $55.9 \pm 8.1$ & $\textbf{80.5} \pm \textbf{2.9}$\\
& pointmaze-giant-navigate-v0   & $4.5 \pm 7.0$  & $0 \pm 0$  & $0 \pm 0$  & $64.7 \pm 13.9$ & $21.1 \pm 10.3$ & $54.7 \pm 10.0$ & $\textbf{74.4} \pm \textbf{5.9}$\\
& pointmaze-teleport-navigate-v0& $26.9 \pm 2.9$ & $45.6 \pm 2.9$ & $23.8 \pm 4.1$ & $4.2 \pm 5.2$ & $26.4 \pm 6.7$ & $12.0 \pm 5.0$ & $\textbf{39.8} \pm\textbf{6.9}$\\
\cmidrule(l){2-9}
& pointmaze-medium-stitch-v0     & $27.6 \pm 14.5$ & $80.6 \pm 1.6$ & $19.9 \pm 11.1$ & $83.3 \pm 6.7$ & $0 \pm 1$ & $79.0 \pm 11.3$ & $\textbf{97.0} \pm \textbf{3.2}$\\
& pointmaze-large-stitch-v0     & $4.0 \pm 8.0$  & $14.0 \pm 7.3$ & $27.8 \pm 2.7$ & $76.7 \pm 8.9$ & $0 \pm 0$ & $14.3 \pm 3.0$ & $\textbf{25.4} \pm \textbf{9.9}$\\
& pointmaze-teleport-stitch-v0  & $26.7 \pm 3.4$ & $41.8 \pm 3.6$ & $27.7 \pm 5.7$ & $13.8 \pm 10.0$  & $3.2 \pm 2.8$ & $30.7 \pm 3.4$ & $\textbf{50.4} \pm \textbf{2.1}$\\
\midrule
\multirow{4}{*}{\textbf{antmaze}}
& antmaze-medium-navigate-v0     & $29 \pm 4$ & $72.9 \pm 6.4$ & $69.7 \pm 5.9$ & $87.0 \pm 2.9$ & $95 \pm 1$ & $\textbf{96.1} \pm \textbf{0.9}$ & $93.4 \pm 1.8$\\
& antmaze-large-navigate-v0      & $23.3 \pm 2.7$ & $20.4 \pm 3.6$ & $36.2 \pm 4.4$ & $78.4 \pm 2.7$ & $84.4 \pm 3.6$ & $\textbf{90.7} \pm \textbf{2.2}$ & $87.1 \pm 3.5$\\
& antmaze-giant-navigate-v0      & $0 \pm 0$  & $0 \pm 0$  & $0 \pm 0$  & $13.0 \pm 2.7$ & $17.7 \pm 5.0$ & $\textbf{70.4} \pm \textbf{2.1}$ & $64.7 \pm 2.7$\\
& antmaze-teleport-navigate-v0   & $26 \pm 3$ & $38.9 \pm 3.4$ & $32.8 \pm 6.5$ & $34.8 \pm 3.8$ & $52.5 \pm 2.6$ & $42 \pm 3$ & $\textbf{48.5} \pm \textbf{2.8}$\\
\midrule
\multirow{2}{*}{\textbf{humanoidmaze}}
& humanoidmaze-medium-navigate-v0 & $8.3 \pm 1.0$ & $27.9 \pm 5.5$ & $27.4 \pm 1.6$ & $10.8 \pm 8.8$ & $57.5 \pm 3.9$ & $\textbf{88.9} \pm \textbf{2.5}$ & $37.1 \pm 10.8$\\
& humanoidmaze-large-navigate-v0  & $1.0 \pm 0.7$ & $1.8 \pm 0.3$ & $2.3 \pm 0.9$ & $6.1 \pm 1.8$ & $17.3 \pm 2.8$ & $\textbf{48.6} \pm \textbf{5.4}$ & $28.8 \pm 7.9$\\
\bottomrule
\end{tabular}
}
\end{table*}

\begin{table*}[t]
\centering
\setlength{\tabcolsep}{6pt}
\renewcommand{\arraystretch}{1.15}
\caption{Results on pixel-based tasks. We evaluate all methods on four sparse-reward visual AntMaze tasks, where observations are high-dimensional pixel inputs and long-horizon planning is required to reach the specified goal. All results report the average success rate and standard deviation over 5 random seeds.}
\label{tab:visualmaze_results}
\resizebox{\textwidth}{!}{
\begin{tabular}{lllcccccc}
\toprule
Environment & Dataset & GCBC & GCIVL & GCIQL & QRL & CRL & HIQL & HIFQL (ours)\\
\midrule
\multirow{4}{*}{\textbf{visual-antmaze}}
& visual-antmaze-medium-navigate-v0   & $6.5 \pm 1.4$ & $14.2 \pm 3.8$ & $5.8 \pm 1.8$ & $0 \pm 0$ & $\textbf{94.3} \pm \textbf{1.3}$ & $87.9 \pm 2.6$ & ${93.3} \pm {2.8}$\\
& visual-antmaze-large-navigate-v0      & $1.4 \pm 1.1$ & $3.1 \pm 1.4$ & $1.8 \pm 0.6$ & $0 \pm 0$ & $\textbf{82.6} \pm \textbf{1.6}$ & $52.9 \pm 5.6$ & ${74.8} \pm {1.9}$\\
& visual-antmaze-giant-navigate-v0      & $0 \pm 0$  & $0.5 \pm 0.3$  & $0.1 \pm 0.2$  & $0 \pm 0$ & $\textbf{32.7} \pm \textbf{4.2}$ & $3.5 \pm 3.0$ & ${30.2} \pm {4.1}$\\
& visual-antmaze-teleport-navigate-v0   & $8.6 \pm 2.3$ & $11.9 \pm 3.1$ & $11.3 \pm 1.0$ & $9.2 \pm 4.0$ & $46.7 \pm 3.5$ & $\textbf{47.0} \pm \textbf{3.8}$ & $43.0 \pm2.6$\\
\bottomrule
\end{tabular}
}
\end{table*}

\textbf{Baselines.} \quad
We compare our method against six representative offline GCRL baselines included in OGBench, covering imitation-based, value-based, metric-based, contrastive, and hierarchical approaches.
\begin{itemize}
    \item Goal-conditioned behavioral cloning (\textbf{GCBC})~\citep{ghosh2019learning} is a simple imitation learning baseline that directly learns a goal-conditioned policy by mimicking actions from the offline dataset.
    \item Goal-conditioned implicit V-learning (\textbf{GCIVL}) and goal-conditioned implicit Q-learning (\textbf{GCIQL})~\citep{kostrikov2021offline, park2023hiql} estimate the goal-conditioned optimal value function using IQL-style expectile regression. Policies are then extracted using advantage-weighted regression (AWR)~\citep{peng2019advantage} for GCIVL and behavior-regularized deep deterministic policy gradient (DDPG+BC)~\citep{fujimoto2021minimalist} for GCIQL, respectively.
    \item Quasimetric RL (\textbf{QRL})~\citep{wang2023optimal} learns a quasimetric value function that estimates the undiscounted temporal distance between states and goals, and derives a policy using DDPG+BC.
    \item Contrastive RL (\textbf{CRL})~\citep{eysenbach2022contrastive} approximates the Q-function via contrastive learning between state–action pairs and future states sampled from the same trajectory, and also extracts a policy using DDPG+BC.
    \item \textbf{HIQL}~\citep{park2023hiql} constructs a two-level hierarchical policy by extracting both a high-level subgoal policy and a low-level action policy from a single GCIVL value function.
\end{itemize}

\subsection{Results on state-based tasks}


We evaluate all methods on three Maze environment families—PointMaze, AntMaze, and HumanoidMaze—covering navigation, stitching, and teleport variants with increasing state dimensionality and planning horizon (Table~\ref{tab:maze_results}).
HIFQL achieves substantial improvements on PointMaze tasks, especially in large-scale and stitching settings where long-horizon planning and multimodal behavior are critical, consistently outperforming HIQL and other baselines. In contrast, on AntMaze and HumanoidMaze tasks, HIFQL does not consistently surpass HIQL, suggesting that when performance is dominated by low-level locomotion and contact-rich dynamics, gains from increased policy expressivity become less pronounced.

\subsection{Results on pixel-based tasks}
Consistent with the state-based results, Table~\ref{tab:visualmaze_results} further evaluates HIFQL in pixel-based Visual AntMaze tasks, where high-dimensional observations place greater demands on goal representation quality. HIFQL achieves strong performance on the medium and large navigation tasks, outperforming HIQL, which highlights the benefit of improved goal representation in visual, long-horizon tasks.
However, on the teleport navigation tasks, HIFQL performs comparably to, but does not surpass HIQL, suggesting reduced gains in the most challenging pixel-based environments.
This is left for future work.

\subsection{Ablation study on representation regularization}
\begin{figure}
    \centering
    \includegraphics[width=1\linewidth]{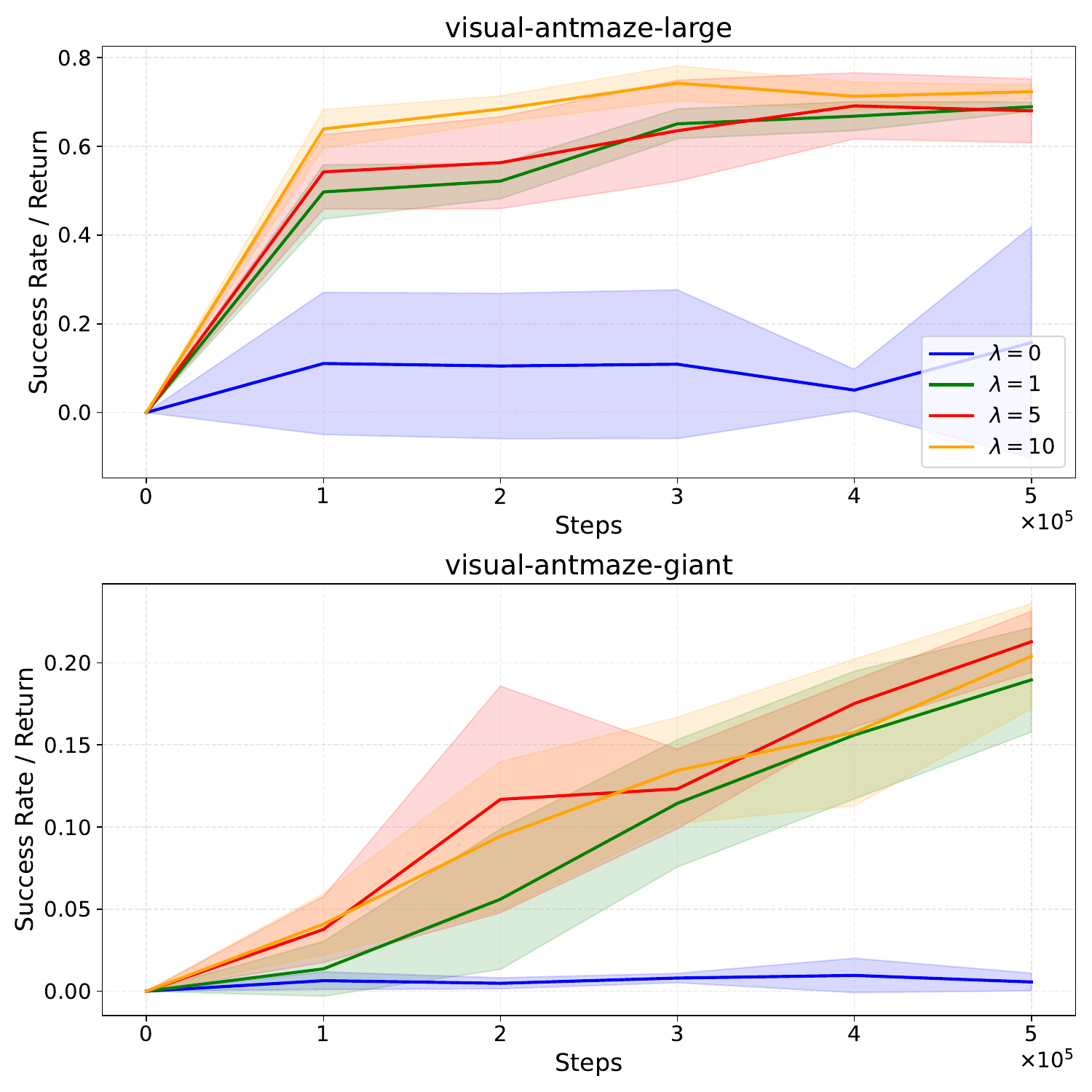}
    \caption{Ablation study on the representation regularization coefficient $\lambda$ in visual AntMaze tasks.}
    \label{fig:lambda}
\end{figure}
To verify the performance of representation regularization, we design ablation experiments with different regularization coefficients $\lambda$.
Figure~\ref{fig:lambda} shows that without goal representation regularization (i.e., $\lambda=0$) leads to consistently low success rates and unstable learning on both visual AntMaze tasks. In contrast, introducing a non-zero $\lambda$ yields substantially higher final performance. This gap highlights that explicit goal representation regularization is critical for effective hierarchical planning in high-dimensional, pixel-based environments.

\section{Conclusion and Discussion}
In this paper, we propose Hierarchical Implicit Flow Q-Learning (HIFQL), an offline goal-conditioned reinforcement learning~(GCRL) method that enhances hierarchical policy learning with expressive mean flow policies and improved goal representations. By replacing unimodal Gaussian policies in HIQL with mean flow models at both hierarchical levels, HIFQL enables efficient one-step inference while better capturing multimodal behaviors. Experiments on OGBench show that HIFQL achieves clear gains on PointMaze tasks, particularly in large-scale and stitching tasks where long-horizon planning and behavioral diversity are critical. On AntMaze and HumanoidMaze tasks, HIFQL does not consistently outperform HIQL and underperforms it in several cases, indicating that when task performance is primarily constrained by low-level locomotion and contact-rich dynamics, gains from increased high-level policy expressivity are limited. These results highlight the effectiveness of flow-based hierarchical policies for long-horizon planning, while also pointing to future opportunities in combining expressive high-level planning with stronger low-level control mechanisms.

\bibliography{reference}

@article{park2024ogbench,
  title={Ogbench: Benchmarking offline goal-conditioned rl},
  author={Park, Seohong and Frans, Kevin and Eysenbach, Benjamin and Levine, Sergey},
  journal={arXiv preprint arXiv:2410.20092},
  year={2024},
  url={https://arxiv.org/pdf/2410.20092}
}

@article{park2023hiql,
  title={Hiql: Offline goal-conditioned rl with latent states as actions},
  author={Park, Seohong and Ghosh, Dibya and Eysenbach, Benjamin and Levine, Sergey},
  journal={Advances in Neural Information Processing Systems},
  volume={36},
  pages={34866--34891},
  year={2023},
  url={}
}

@article{park2025flow,
  title={Flow q-learning},
  author={Park, Seohong and Li, Qiyang and Levine, Sergey},
  journal={arXiv preprint arXiv:2502.02538},
  year={2025}
}

@article{lipman2022flow,
  title={Flow matching for generative modeling},
  author={Lipman, Yaron and Chen, Ricky TQ and Ben-Hamu, Heli and Nickel, Maximilian and Le, Matt},
  journal={arXiv preprint arXiv:2210.02747},
  year={2022}
}

@article{kostrikov2021offline,
  title={Offline reinforcement learning with implicit q-learning},
  author={Kostrikov, Ilya and Nair, Ashvin and Levine, Sergey},
  journal={arXiv preprint arXiv:2110.06169},
  year={2021}
}

@article{peng2019advantage,
  title={Advantage-weighted regression: Simple and scalable off-policy reinforcement learning},
  author={Peng, Xue Bin and Kumar, Aviral and Zhang, Grace and Levine, Sergey},
  journal={arXiv preprint arXiv:1910.00177},
  year={2019}
}

@article{kang2023efficient,
  title={Efficient diffusion policies for offline reinforcement learning},
  author={Kang, Bingyi and Ma, Xiao and Du, Chao and Pang, Tianyu and Yan, Shuicheng},
  journal={Advances in Neural Information Processing Systems},
  volume={36},
  pages={67195--67212},
  year={2023}
}

@article{ghosh2019learning,
  title={Learning to reach goals via iterated supervised learning},
  author={Ghosh, Dibya and Gupta, Abhishek and Reddy, Ashwin and Fu, Justin and Devin, Coline and Eysenbach, Benjamin and Levine, Sergey},
  journal={arXiv preprint arXiv:1912.06088},
  year={2019}
}

@article{eysenbach2022contrastive,
  title={Contrastive learning as goal-conditioned reinforcement learning},
  author={Eysenbach, Benjamin and Zhang, Tianjun and Levine, Sergey and Salakhutdinov, Russ R},
  journal={Advances in Neural Information Processing Systems},
  volume={35},
  pages={35603--35620},
  year={2022}
}

@article{ahn2025option,
  title={Option-aware Temporally Abstracted Value for Offline Goal-Conditioned Reinforcement Learning},
  author={Ahn, Hongjoon and Choi, Heewoong and Han, Jisu and Moon, Taesup},
  journal={arXiv preprint arXiv:2505.12737},
  year={2025}
}

@article{lipman2024flow,
  title={Flow matching guide and code},
  author={Lipman, Yaron and Havasi, Marton and Holderrieth, Peter and Shaul, Neta and Le, Matt and Karrer, Brian and Chen, Ricky TQ and Lopez-Paz, David and Ben-Hamu, Heli and Gat, Itai},
  journal={arXiv preprint arXiv:2412.06264},
  year={2024}
}

@article{geng2025mean,
  title={Mean flows for one-step generative modeling},
  author={Geng, Zhengyang and Deng, Mingyang and Bai, Xingjian and Kolter, J Zico and He, Kaiming},
  journal={arXiv preprint arXiv:2505.13447},
  year={2025}
}

@article{huang2019mapping,
  title={Mapping state space using landmarks for universal goal reaching},
  author={Huang, Zhiao and Liu, Fangchen and Su, Hao},
  journal={Advances in Neural Information Processing Systems},
  volume={32},
  year={2019}
}

@article{kim2021landmark,
  title={Landmark-guided subgoal generation in hierarchical reinforcement learning},
  author={Kim, Junsu and Seo, Younggyo and Shin, Jinwoo},
  journal={Advances in neural information processing systems},
  volume={34},
  pages={28336--28349},
  year={2021}
}

@article{liu2022flow,
  title={Flow straight and fast: Learning to generate and transfer data with rectified flow},
  author={Liu, Xingchao and Gong, Chengyue and Liu, Qiang},
  journal={arXiv preprint arXiv:2209.03003},
  year={2022}
}

@article{albergo2022building,
  title={Building normalizing flows with stochastic interpolants},
  author={Albergo, Michael S and Vanden-Eijnden, Eric},
  journal={arXiv preprint arXiv:2209.15571},
  year={2022}
}

@article{ho2020denoising,
  title={Denoising diffusion probabilistic models},
  author={Ho, Jonathan and Jain, Ajay and Abbeel, Pieter},
  journal={Advances in neural information processing systems},
  volume={33},
  pages={6840--6851},
  year={2020}
}

@inproceedings{kaelbling1993learning,
  title={Learning to achieve goals},
  author={Kaelbling, Leslie Pack},
  booktitle={IJCAI},
  volume={2},
  pages={1094--8},
  year={1993}
}

@article{liu2022goal,
  title={Goal-conditioned reinforcement learning: Problems and solutions},
  author={Liu, Minghuan and Zhu, Menghui and Zhang, Weinan},
  journal={arXiv preprint arXiv:2201.08299},
  year={2022}
}

@inproceedings{wang2023optimal,
  title={Optimal goal-reaching reinforcement learning via quasimetric learning},
  author={Wang, Tongzhou and Torralba, Antonio and Isola, Phillip and Zhang, Amy},
  booktitle={International Conference on Machine Learning},
  pages={36411--36430},
  year={2023},
  organization={PMLR}
}

@article{myers2024learning,
  title={Learning temporal distances: Contrastive successor features can provide a metric structure for decision-making},
  author={Myers, Vivek and Zheng, Chongyi and Dragan, Anca and Levine, Sergey and Eysenbach, Benjamin},
  journal={arXiv preprint arXiv:2406.17098},
  year={2024}
}

@article{ma2022vip,
  title={Vip: Towards universal visual reward and representation via value-implicit pre-training},
  author={Ma, Yecheng Jason and Sodhani, Shagun and Jayaraman, Dinesh and Bastani, Osbert and Kumar, Vikash and Zhang, Amy},
  journal={arXiv preprint arXiv:2210.00030},
  year={2022}
}

@article{liu2024single,
  title={A single goal is all you need: Skills and exploration emerge from contrastive rl without rewards, demonstrations, or subgoals},
  author={Liu, Grace and Tang, Michael and Eysenbach, Benjamin},
  journal={arXiv preprint arXiv:2408.05804},
  year={2024}
}

@inproceedings{sohl2015deep,
  title={Deep unsupervised learning using nonequilibrium thermodynamics},
  author={Sohl-Dickstein, Jascha and Weiss, Eric and Maheswaranathan, Niru and Ganguli, Surya},
  booktitle={International conference on machine learning},
  pages={2256--2265},
  year={2015},
  organization={pmlr}
}

@article{dhariwal2021diffusion,
  title={Diffusion models beat gans on image synthesis},
  author={Dhariwal, Prafulla and Nichol, Alexander},
  journal={Advances in neural information processing systems},
  volume={34},
  pages={8780--8794},
  year={2021}
}

@inproceedings{esser2024scaling,
  title={Scaling rectified flow transformers for high-resolution image synthesis},
  author={Esser, Patrick and Kulal, Sumith and Blattmann, Andreas and Entezari, Rahim and M{\"u}ller, Jonas and Saini, Harry and Levi, Yam and Lorenz, Dominik and Sauer, Axel and Boesel, Frederic and others},
  booktitle={Forty-first international conference on machine learning},
  year={2024}
}

@article{janner2022planning,
  title={Planning with diffusion for flexible behavior synthesis},
  author={Janner, Michael and Du, Yilun and Tenenbaum, Joshua B and Levine, Sergey},
  journal={arXiv preprint arXiv:2205.09991},
  year={2022}
}

@article{ajay2022conditional,
  title={Is conditional generative modeling all you need for decision-making?},
  author={Ajay, Anurag and Du, Yilun and Gupta, Abhi and Tenenbaum, Joshua and Jaakkola, Tommi and Agrawal, Pulkit},
  journal={arXiv preprint arXiv:2211.15657},
  year={2022}
}

@article{liang2023adaptdiffuser,
  title={Adaptdiffuser: Diffusion models as adaptive self-evolving planners},
  author={Liang, Zhixuan and Mu, Yao and Ding, Mingyu and Ni, Fei and Tomizuka, Masayoshi and Luo, Ping},
  journal={arXiv preprint arXiv:2302.01877},
  year={2023}
}

@article{wang2022diffusion,
  title={Diffusion policies as an expressive policy class for offline reinforcement learning},
  author={Wang, Zhendong and Hunt, Jonathan J and Zhou, Mingyuan},
  journal={arXiv preprint arXiv:2208.06193},
  year={2022}
}

@article{lu2024simplifying,
  title={Simplifying, stabilizing and scaling continuous-time consistency models},
  author={Lu, Cheng and Song, Yang},
  journal={arXiv preprint arXiv:2410.11081},
  year={2024}
}

@article{song2023consistency,
  title={Consistency models},
  author={Song, Yang and Dhariwal, Prafulla and Chen, Mark and Sutskever, Ilya},
  year={2023}
}

@article{geng2024consistency,
  title={Consistency models made easy},
  author={Geng, Zhengyang and Pokle, Ashwini and Luo, William and Lin, Justin and Kolter, J Zico},
  journal={arXiv preprint arXiv:2406.14548},
  year={2024}
}

@article{levine2020offline,
  title={Offline reinforcement learning: Tutorial, review, and perspectives on open problems},
  author={Levine, Sergey and Kumar, Aviral and Tucker, George and Fu, Justin},
  journal={arXiv preprint arXiv:2005.01643},
  year={2020}
}

@article{pateria2021hierarchical,
  title={Hierarchical reinforcement learning: A comprehensive survey},
  author={Pateria, Shubham and Subagdja, Budhitama and Tan, Ah-hwee and Quek, Chai},
  journal={ACM Computing Surveys (CSUR)},
  volume={54},
  number={5},
  pages={1--35},
  year={2021},
  publisher={ACM New York, NY, USA}
}

@article{mandlekar2021matters,
  title={What matters in learning from offline human demonstrations for robot manipulation},
  author={Mandlekar, Ajay and Xu, Danfei and Wong, Josiah and Nasiriany, Soroush and Wang, Chen and Kulkarni, Rohun and Fei-Fei, Li and Savarese, Silvio and Zhu, Yuke and Mart{\'\i}n-Mart{\'\i}n, Roberto},
  journal={arXiv preprint arXiv:2108.03298},
  year={2021}
}

@inproceedings{o2024open,
  title={Open x-embodiment: Robotic learning datasets and rt-x models: Open x-embodiment collaboration 0},
  author={O’Neill, Abby and Rehman, Abdul and Maddukuri, Abhiram and Gupta, Abhishek and Padalkar, Abhishek and Lee, Abraham and Pooley, Acorn and Gupta, Agrim and Mandlekar, Ajay and Jain, Ajinkya and others},
  booktitle={2024 IEEE International Conference on Robotics and Automation (ICRA)},
  pages={6892--6903},
  year={2024},
  organization={IEEE}
}

@article{chi2025diffusion,
  title={Diffusion policy: Visuomotor policy learning via action diffusion},
  author={Chi, Cheng and Xu, Zhenjia and Feng, Siyuan and Cousineau, Eric and Du, Yilun and Burchfiel, Benjamin and Tedrake, Russ and Song, Shuran},
  journal={The International Journal of Robotics Research},
  volume={44},
  number={10-11},
  pages={1684--1704},
  year={2025},
  publisher={Sage Publications Sage UK: London, England}
}

@article{fujimoto2021minimalist,
  title={A minimalist approach to offline reinforcement learning},
  author={Fujimoto, Scott and Gu, Shixiang Shane},
  journal={Advances in neural information processing systems},
  volume={34},
  pages={20132--20145},
  year={2021}
}

@article{balestriero2025lejepa,
  title={Lejepa: Provable and scalable self-supervised learning without the heuristics},
  author={Balestriero, Randall and LeCun, Yann},
  journal={arXiv preprint arXiv:2511.08544},
  year={2025}
}

@article{monemi2025tutorial,
  title={Tutorial on Joint Embedding Predictive Architectures (JEPA): Foundations, Applications, and Future Directions},
  author={Monemi, Mehdi and Chinipardaz, Maryam and Rasti, Mehdi and Bennis, Mehdi and Latva-Aho, Matti},
  journal={Authorea Preprints},
  year={2025},
  publisher={Authorea}
}

@article{bromley1993signature,
  title={Signature verification using a" siamese" time delay neural network},
  author={Bromley, Jane and Guyon, Isabelle and LeCun, Yann and S{\"a}ckinger, Eduard and Shah, Roopak},
  journal={Advances in neural information processing systems},
  volume={6},
  year={1993}
}

@article{lecun2022path,
  title={A path towards autonomous machine intelligence version 0.9. 2, 2022-06-27},
  author={LeCun, Yann},
  journal={Open Review},
  volume={62},
  number={1},
  pages={1--62},
  year={2022}
}

@inproceedings{chen2020simple,
  title={A simple framework for contrastive learning of visual representations},
  author={Chen, Ting and Kornblith, Simon and Norouzi, Mohammad and Hinton, Geoffrey},
  booktitle={International conference on machine learning},
  pages={1597--1607},
  year={2020},
  organization={PmLR}
}

@inproceedings{chen2021empirical,
  title={An empirical study of training self-supervised vision transformers},
  author={Chen, Xinlei and Xie, Saining and He, Kaiming},
  booktitle={Proceedings of the IEEE/CVF international conference on computer vision},
  pages={9640--9649},
  year={2021}
}
\bibliographystyle{unsrt}

\end{document}